\documentclass{article}

\usepackage[preprint]{neurips_2026}

\usepackage[utf8]{inputenc} 
\usepackage[T1]{fontenc}    
\usepackage{hyperref}       
\usepackage{url}            
\usepackage{booktabs}       
\usepackage{amsfonts}       
\usepackage{nicefrac}       
\usepackage{microtype}      
\usepackage{xcolor}         
\usepackage{amsmath}
\usepackage{multirow}
\usepackage{array}
\usepackage{graphicx}

\title{When2Speak: A Dataset for Temporal Participation and Turn-Taking in Multi-Party Conversations for Large Language Models}

\author{%
  Vihaan Nama \\
  Pratt School of Engineering\\
  Duke University\\
  \texttt{vihaan.nama@outlook.com} \\
  \And
  Shreya Mendi \\
  Pratt School of Engineering\\
  Duke University\\
  \texttt{shreya.mendi@duke.edu} \\
  \And
  Zian Ye \\
  Pratt School of Engineering\\
  Duke University\\
  \texttt{zy223@duke.edu} \\
  \And
  Dr.\ Brinnae Bent \\
  Pratt School of Engineering\\
  Duke University\\
  \texttt{brinnae.bent@duke.edu} \\
}

\begin{document}

\maketitle

\begin{abstract}
Large Language Models (LLMs) excel at generating contextually appropriate responses but remain poorly calibrated for multi-party conversations, where deciding when to speak is as critical as what to say. In such settings, naïvely responding at every turn leads to excessive interruptions and degraded conversational coherence. We introduce When2Speak, a grounded synthetic dataset and four-stage generation pipeline for learning intervention timing in group interactions. The dataset comprises over 215,000 examples derived from 16,000 conversations involving 2–6 speakers, spanning diverse conversational styles, tones, and participant dynamics, and explicitly modeling SPEAK vs. SILENT decisions at each turn. Our pipeline combines real-world grounding, structured augmentation, controlled transcript synthesis, and fine-tuning-ready supervision, and is fully open-sourced to support reproducibility and adaptation to domain-specific conversational norms. Across multiple model families, supervised fine-tuning (SFT) on When2Speak significantly outperforms zero-shot baselines (e.g., the average Macro F1 increase across 4B+ parameter models was 60\%, with the largest increase being 120\%). However, SFT-trained models remain systematically over-conservative, missing nearly half of warranted interventions as seen through the Missed Intervention Rate (MIR), which was on average 0.50 and is noticed even at larger model sizes. To address this limitation, we apply reinforcement learning with asymmetric reward shaping, which reduces MIR to 0.186–0.218 and increases recall from 0.479 to 0.78–0.81. Our findings establish that temporal participation is a distinct and trainable dimension of conversational intelligence, and that grounded synthetic data provides an effective and scalable pathway for enabling LLMs to participate more naturally and appropriately in multi-party interactions.

\end{abstract}


\section{Introduction}

Large Language Models (LLMs) excel at generating high-quality responses, but their behavior in multi-party conversations remains fundamentally limited by an underexplored capability: deciding when to speak. Existing training paradigms optimize models to produce the best possible response conditioned on a prompt, implicitly assuming that a response is always warranted. In real-world group interactions, conversational quality depends equally on the ability to withhold unnecessary interventions and participate only at contextually appropriate moments, as current dialogue systems often fail to determine when to speak and may interrupt excessively \citep{arora2025}. When this temporal participation signal is absent, LLMs tend to over-interject \citep{umair2024}, disrupting established turn-taking dynamics in human conversation \citep{sacks1974}.
This limitation persists largely because current datasets and benchmarks emphasize response generation quality rather than participation timing, leaving the problem of deciding when to intervene unrepresented in existing training pipelines. 

To address this gap, we introduce When2Speak, a grounded synthetic dataset and modular generation pipeline for teaching LLMs when to intervene and when to remain silent in ongoing multi-speaker interactions. Our pipeline combines real-world grounding, transcript generation, silence modeling, and fine-tuning supervision to learn when an LLM should speak.
To validate the effectiveness of the dataset, we perform both supervised fine-tuning (SFT) and reinforcement learning (RL) across multiple model families, including GPT-4.1-mini, Qwen3 (4B, 8B), Llama-3.1 (8B), and Llama-3.3 (70B). Beyond the dataset itself, our work establishes a reproducible and open-sourced framework for grounded synthetic data generation and supervision of conversational timing behaviors, highlighting a broader pathway for extending LLM alignment beyond response correctness toward interaction-level conversational intelligence.

\section{Related Works}

Research on conversational turn-taking originates with \citet{sacks1974}, which introduced Transition Relevance Places as moments when speaker change becomes relevant. Computational dialogue systems have since modeled turn-taking in spoken and text-based settings, often by predicting when a user has finished speaking \citep{skantze2021,ekstedt2020,jiang2023}. However, recent work shows that even strong LLMs struggle to identify transition-relevant places in naturalistic dialogue \citep{umair2024}. Additionally, these studies primarily focus on interactions between two parties, which may not extend to multi-party conversations.

A related line of literature studies LLM abstention, where models learn to withhold answers under uncertainty, insufficient evidence, or hallucination risk \citep{zhang2024,feng2024,wen2025}. While this establishes silence as a desirable model behavior, it frames non-response mainly as an epistemic choice. In multi-party dialogue, silence is also interactional: an agent may know a useful answer but still need to remain silent because an intervention would interrupt the flow, duplicate another participant's role, or reduce conversational coherence.

Dataset availability further limits progress on this problem. Existing multi-party corpora include meetings, scripted dialogue, online discussions, and earnings calls \citep{carletta2006,chen2016,ouchi2016,oneill2021}, but they were not collected with an AI agent present in the environment. \citet{wei2023} move closer to this setting with MultiLIGHT, a crowdsourced multi-party role-play dataset for training conversational agents to decide when to speak and generate utterances grounded in multiple participants. However, MultiLIGHT models agents as role-playing characters in a symmetric group interaction, rather than as an assistant deciding whether its contribution is warranted. Speak or Stay Silent \citep{bhagtani2026} takes a different approach by repurposing existing real-world multi-party corpora: they identify candidate turn-transition points in the AMI, Friends, and SPGISpeech datasets, then relabel each context as a binary speak-or-stay-silent decision for an LLM. This creates a large-scale benchmark from naturally occurring interaction patterns and shows that context-aware turn-taking requires explicit training.

Synthetic dialogue generation offers a complementary approach to creating supervision when naturally occurring data is limited. Prior work has shown that synthetic conversations can support dialogue training through persona-conditioned generation, self-chat, commonsense grounding, and instruction-style supervision \citep{kim2023,ding2023,xu2023,suresh2025}. DiscussLLM applies this strategy to the when-to-speak problem by generating conversations with silent-token supervision and intervention types such as factual correction, concept definition, data provision, source identification, and synthesis \citep{patel2025}. Even though this approach was a step in the right direction, it lacked an open-sourced dataset and framework for reproducibility, and it did not account for various tones, styles, follow-up questions, and variations of the conversation. When2Speak builds on this direction with grounded synthetic multi-party conversations with a present AI agent, explicit SPEAK/SILENT decisions, controlled variation in tone and interaction style, and post-intervention follow-up dynamics.

\begin{table}[t]
\caption{Comparison between When2Speak and prior work on multi-party intervention timing datasets. When2Speak is the only approach that is fully open-sourced, includes RL training, and covers all five design dimensions.}
\label{tab:comparison}
\centering
\resizebox{\linewidth}{!}{%
\begin{tabular}{lccccccccr}
\hline
 & \textbf{Tone} & \textbf{Conversational} & \textbf{Follow-Up} & \textbf{Intervention} & \textbf{Open} & \textbf{Open} & & & \textbf{Num.} \\
 & \textbf{Variations Types} & \textbf{Styles Types} & \textbf{Question Types} & \textbf{Types} & \textbf{Data} & \textbf{Code} & \textbf{SFT} & \textbf{RL} & \textbf{Examples} \\
\hline
When2Speak           & 5 & 6 & 4 & 5 & $\checkmark$ & $\checkmark$ & $\checkmark$ & $\checkmark$ & 216k \\
Speak or Stay Silent & N/A & N/A  & N/A     & N/A     & $\checkmark$ & $\checkmark$ & $\checkmark$ & $\times$ & 120k \\
DiscussLLM           & N/A     & N/A     & N/A     & 5 & $\times$ & $\times$ & $\checkmark$ & $\times$ & N/A \\
\hline
\end{tabular}}
\end{table}

\section{When2Speak}

To construct a high-quality dataset for modeling intervention timing in multi-party conversations, we develop a multi-stage generation pipeline that integrates grounded real-world data with controlled synthetic transcript generation. The pipeline begins with a diverse source corpus, followed by structured augmentation to introduce social context and intervention types, and a transcript synthesis stage that captures realistic conversational dynamics. The generated conversations are then transformed into fine-tuning-ready training examples using a sliding-window formulation and serialized into a JSONL format. We adapt approaches from DiscussLLM \citep{patel2025}; specifically, dataset grounding, social context and intervention identification, and tokenization of SILENT. We contribute by adding tone and style variations and introducing follow-up questions. This process is extremely important, as it models how real-world conversations are structured, a feature lacking in other related works. This process yields a dataset of 216,799 examples, derived from 16,000 distinct conversations, designed to model when an LLM should intervene or remain silent. All synthetic data is generated using GPT-4o-mini (Stage 1) and GPT-4-turbo-preview (Stage 2) via structured prompting. To build this dataset, the help of AWS services were used, specifically - AWS EC2-c6i.4xlarge, consisting of 8 cores and 16 vCPUs. The following sections provide a detailed description of each stage in the pipeline.

\subsection{Stage 1: Grounded Source Dataset Selection and Cleaning}

Stage 1 establishes the grounded source corpus for all subsequent synthetic data generation. Since synthetic data quality heavily depends on grounding in a diverse real-world distribution, we select the Yahoo Answers dataset from Kaggle \citep{ardeshna2018}, originally used as a text-classification benchmark by \citet{zhang2015}. We use a publicly available dataset released under a CC0 (public domain) license for distributional grounding during scenario synthesis. All final dataset instances are generated de novo using language models. The released dataset does not contain or reconstruct any original samples from the source corpus. 

We retain three fields from each example: Question Title, Question Content, and Best Answer, which together provide a diverse and well-structured foundation for questions and responses. We then apply a cleaning pipeline by removing rows with missing values, discarding examples where any retained field contains fewer than 5 words, and filtering rows containing special characters. After cleaning, we select the top 16,000 rows as the grounded truth for the source conversations derived in the following stages and used throughout the remainder of the pipeline. 

\subsection{Stage 2: Social Context and Intervention Type Augmentation}

The second stage of the pipeline focuses on enriching the cleaned Yahoo Answers dataset with two additional attributes: social context and intervention type. For each example, we provide an LLM with the question title, question content, and best answer, all sourced from the cleaned Yahoo Answers dataset. Conditioned on these inputs, the model infers 1.\ \textit{a social context}, defined as a single-sentence description of the type of individuals who would plausibly engage in a discussion on the given topic, and 2.\ an \textit{intervention type}, corresponding to the most appropriate \texttt{ai\_intervention\_type} selected from the predefined set: [\textit{Factual Correction, Concept Definition, Data Provision, Source Identification, Synthesis \& Reframing}] \citep{patel2025}. This prompting procedure (Appendix ~\ref{app:stage2-prompt}) is executed row-wise across the entire cleaned dataset, resulting in an augmented version in which each original example is paired with inferred conversational metadata.

We use GPT-4o-mini to infer social context and intervention type via structured prompting.

\subsection{Stage 3: Controlled Multi-Party Transcript Synthesis}
In this stage, we synthesize multi-party conversational transcripts for each row in the Stage 2 augmented dataset. We prompt the LLM to operate as a sophisticated data generator and generate a transcript conditioned on the structured metadata produced in Stage 2 (Appendix ~\ref{app:stage3-prompt}). Specifically, each prompt includes the topic, initial question, social context, AI intervention type, and reference answer (included as additional grounding context). The objective of this stage is to transform static question--answer examples into realistic, turn-by-turn group conversations that explicitly model both human interaction dynamics and appropriate LLM participation.
 
A key challenge in transcript synthesis is ensuring coverage across the diverse formats that real-world conversations can take. To address this, we define a curated set of conversation styles that specify the desired discourse structure for each generated transcript. These styles include `debates with frequent disagreement', `collaborative discussions where participants build on each other's ideas', `interactions between experts and novices involving clarifying questions', `storytelling with personal anecdotes', `structured conversations with formal turn-taking', and `casual dialogues characterized by overlap and interruptions'. We incorporate style control to enable the dataset to capture a broader spectrum of interaction patterns while improving standardization and real-world fidelity. In parallel, we model conversational tone, recognizing that tone can materially alter the interpretation and flow of dialogue. We therefore define a list of tonality variations, spanning `enthusiastic and energetic', `thoughtful and contemplative', `professional and business-like', `casual and friendly', and `curious and inquisitive'. For every generated transcript, we randomly sample one conversation style and one tone, thereby introducing controlled diversity across the dataset.
 
Because multi-party conversations often continue after an LLM intervention, we additionally model post-intervention follow-up scenarios that reflect plausible human responses to agent participation. To capture this, we construct a predefined set of follow-up interaction types, including: (1) participants asking the agent a follow-up question and receiving a response, (2) participants refuting the agent's contribution and requiring a deeper explanation, (3) the group agreeing with the intervention and continuing the discussion, and (4) the group disagreeing and moving forward without further engagement. Since the LLM may intervene at multiple points in a conversation, including these multi-turn follow-up branches is critical for ensuring that the dataset captures realistic downstream conversational trajectories and edge cases. For each transcript, one follow-up type is randomly selected from this predefined set.
 
We further impose a set of explicit generation constraints that improve transcript quality and consistency. The LLM is instructed to ensure that the agent appears between 1 and 3 times per conversation, with each intervention arising from a natural and clearly identifiable conversational trigger. Each transcript must contain at least 20 exchanges, and every participant is required to exhibit a distinct conversational personality. To vary interaction complexity, the number of human speakers is randomly sampled between 2 and 6 participants for each transcript. All participants are normalized to anonymized identifiers such as \texttt{Speaker\_0}, \texttt{Speaker\_1}, \ldots, reducing the risk of overfitting to specific names or identity cues. Participants are also instructed to treat the agent as an AI conversational partner, allowing both explicit direct questions and implicit invitations for intervention.
 
The silent token is defined as a token with no semantic meaning in natural language, allowing it to represent intentional silence \citep{patel2025}. This mechanism is particularly important because, at every conversational turn, the LLM must emit an output, even when the correct behavior is to remain silent. To model this behavior, we introduce a dedicated silence token and select `\texttt{>}' for this purpose, as it carries no standalone semantic meaning in English discourse and does not function as conventional punctuation in this setting. Whenever the agent should remain silent, the transcript records the \texttt{>} token; whenever intervention is appropriate, the transcript instead contains the corresponding natural-language response.
 
The final output of this stage is a collection of diverse, grounded multi-party transcripts involving 2--6 speakers, each containing at least 20 exchanges, varied tones, and multiple conversational styles. An LLM agent is explicitly embedded within each conversation, emitting either the silent token \texttt{>} when silence is appropriate or the corresponding intervention utterance when participation is required. The transcripts additionally incorporate realistic post-intervention follow-up trajectories, resulting in a richly structured dataset that is ready for downstream training and evaluation.

Transcripts are generated using GPT-4-turbo-preview conditioned on structured inputs and generation constraints.

\subsection{Stage 4: Dataset Variant Construction}
We convert the raw transcript outputs produced in Stage 3 into a structured training format suitable for model consumption and downstream fine-tuning. An 8-message sliding-window within a conversation was used as context, with the label being an agent's decision to respond or remain silent at each turn.
 
We then construct two parallel dataset variants to evaluate whether explicit speech-action tokenization improves learnability. In realistic conversations, the LLM should remain silent for the majority of turns, while intervention turns occur under highly varied conversational conditions and may contain semantically diverse responses \citep{patel2025, bhagtani2026}. This creates a highly imbalanced and sparse dataset, which may increase the risk of the model overfitting to the silent token behavior. To study this effect, we construct two parallel dataset variants.
 
The first variant, referred to as the \textit{Token Dataset}, replaces every agent intervention utterance with a dedicated \texttt{<} speak token, thereby collapsing all intervention behaviors into a unified action representation. The second variant, referred to as the \textit{Dialogue Dataset}, preserves the agent's original natural-language intervention text exactly as generated in Stage 2. The \textit{Token Dataset} is intended to reduce output-space complexity and mitigate the possibility of overfitting towards silence by reframing the task as a more explicit speak-versus-silent action prediction problem, while the \textit{Dialogue Dataset} retains the full expressive intervention space.
 
Both dataset variants are then serialized into JSONL records, where each line corresponds to a single sliding-window training example and its associated supervision target. This representation ensures direct compatibility with modern LLM supervised fine-tuning frameworks and enables efficient large-scale training.
 
For both the \textit{Token Dataset} and the \textit{Dialogue Dataset}, we generate an 80/10/10 train--test--validation split. It was ensured that this was split on the conversation level so as not to have data leaks between sets. As a result, the final output of this stage consists of six distinct JSONL files: train, validation, and test splits for each of the two dataset variants. The final datasets (Token and Dialogue) each contain 173,325 unique examples (train), 21,687 unique examples (test), and 21,787 unique examples (validation) after some cleaning, resulting in a total dataset size of 216,799 unique examples.

\section{Dataset Statistics}

This section characterizes the scale, structure, and distributional properties of the When2Speak dataset to assess its suitability for training and evaluating intervention policies in grounded multi-party conversations. The analysis spans conversational structure, intervention type distribution, and example-level statistics across the final 216,799-example corpus derived from 16,000 grounded conversations. Through summary statistics and distributional breakdowns, we highlight properties critical to training intervention timing models, including class imbalance, silence prevalence, and intervention diversity.

\subsection{Conversational Structure Analysis}

Across the corpus, the assistant produces a mean of 1.86 interventions per conversation (median $= 2$, std $= 1.2$). Notably, 86.2\% of \textsc{speak} turns ($n = 25{,}621$) occur without the agent being explicitly addressed, requiring the model to infer the appropriateness of intervention purely from the conversational context. In contrast, only 13.8\% ($n = 4{,}089$) of interventions are directly solicited by name. This distribution reflects realistic deployment conditions, where contextual reasoning rather than explicit prompts predominantly drive effective interventions \citep{patel2025}.
 
Importantly, 2.8\% of conversations ($n = 453$) contain zero interventions, with the assistant remaining entirely silent throughout. These examples are critical for learning the \textsc{silent} decision in well-functioning conversations and are preserved without augmentation. Additionally, 17.0\% of conversations ($n = 2{,}713$) contain disagreement, identified via lexical markers.

\subsection{Intervention Type Distribution}

The distribution of intervention types is non-uniform, reflecting natural patterns in group discussions. Synthesis and Reframing is the most frequent category, comprising 5,718 instances (35.7\%), where the agent integrates multiple viewpoints into a coherent perspective. This is followed by Data Provision with 3,903 instances (24.4\%), where the agent contributes relevant factual information. Factual Correction accounts for 2,758 instances (17.2\%), involving the identification and correction of erroneous claims. Source Identification appears in 2,132 instances (13.3\%), where the agent directs participants to authoritative sources for verification. Finally, Concept Definition is the least frequent, with 1,489 instances (9.3\%), focusing on clarifying misunderstood or undefined concepts.

\section{What When2Speak Enables: Learning Intervention Timing}

We leverage the When2Speak dataset to investigate whether language models can learn principled intervention timing. We evaluate two training paradigms: supervised fine-tuning, which establishes a strong baseline for the timing decision, and reinforcement learning with asymmetric reward shaping, which directly optimizes for the cost asymmetry between false interruptions and missed interventions.

\subsection{Supervised Fine-Tuning}

\subsubsection{Experimental Setup}

We fine-tune six open-source LLMs (Llama-3.2-1B, Llama-3.2-3B, Qwen3-4B, Qwen3-8B, Llama-3.1-8B, and Llama-3.3-70B) and two commercial model variants (GPT-4.1-mini, GPT-4o-mini) on the When2Speak \textit{Token} and \textit{Dialogue} Datasets. The open-source models were fine-tuned using LoRA with a rank of 16, a scaling factor $\alpha$ of 32, a learning rate of $1\times10^{-4}$, a batch size of 256, and for 3 epochs. Rank 16 follows the standard LoRA configuration, balancing adapter expressivity with parameter efficiency. The scaling factor $\alpha=32$ ($\alpha/r=2$) is the widely adopted default that scales adapter updates by $2\times$, providing stable training without over-amplifying low-rank gradients. A learning rate of $10^{-4}$ is standard for LoRA fine-tuning, as higher values risk destabilizing pretrained backbone weights. The open-source models were trained using Tinker Labs A100 infrastructure, while the commercial models were trained via the OpenAI fine-tuning API using 3 epochs and a learning rate multiplier of 0.1 with similar remaining parameters. All models are compared against the zero-shot versions of models prompted to output a single token---`\texttt{<}' to speak or `\texttt{>}' to remain silent, given the conversation history (complete prompt in Appendix~\ref{app:stage2-prompt}).

\subsubsection{Evaluation Metrics}

Standard classification metrics are insufficient for evaluating intervention timing, as they either conflate distinct error types or fail to account for class imbalance. In this setting, over-intervention and under-intervention carry fundamentally different costs, requiring more targeted evaluation. We therefore adopt a three-metric evaluation protocol that separates these failure modes and provides a more interpretable assessment of conversational behavior. Our framework captures both the social cost of unnecessary interruptions and the utility cost of missed interventions, enabling a more nuanced analysis of model performance in multi-party settings.

Let TP, TN, FP, FN denote true positives (correct SPEAK), true negatives (correct SILENT), false positives (wrong SPEAK $=$ interruption), and false negatives (wrong SILENT $=$ missed intervention). We adopt Macro F1 as the primary metric and introduce two task-specific metrics:
\begin{align}
\text{FIR} &= \frac{\text{FP}}{\text{FP} + \text{TN}} \quad \text{(False Interruption Rate)} \\[4pt]
\text{MIR} &= \frac{\text{FN}}{\text{FN} + \text{TP}} \quad \text{(Missed Intervention Rate)}
\end{align}

\subsubsection{Results}

Fine-tuning on When2Speak consistently outperforms zero-shot prompting across all models (Table~\ref{tab:sft}). The best SFT model (Llama-3.3-70B) achieves Macro F1 $= 0.747$, exceeding its zero-shot baseline by $+0.422$. Notably, an 8B fine-tuned model (0.740) outperforms GPT-4o (0.567) in zero-shot, indicating that intervention timing is primarily a learned behavior rather than a function of model scale.

Zero-shot models exhibit high recall (0.52--0.69) but suffer from elevated FIR (0.26--0.59), reflecting over-intervention. In contrast, SFT models significantly reduce FIR (0.036--0.050) but incur higher MIR (0.488--0.549), indicating conservative behavior that avoids unnecessary interruptions at the cost of missed interventions.

Two small models (1B, 3B) collapse to the Always-SILENT baseline, suggesting that sub-4B models cannot overcome class imbalance under standard SFT. For non-collapsed models, performance improves with scale: Macro F1 4B $(0.737) \to$ 8B $(0.739$--$0.740) \to$ 70B $(0.747)$, with corresponding decreases in MIR. However, gains beyond 8B are marginal ($+0.006$ F1), and all SFT models remain systematically over-conservative, with the best model missing 48.8\% of required interventions.

Comparing zero-shot and fine-tuned versions of the same models highlights the impact of domain-specific training. In zero-shot, open-source models perform poorly (e.g., Macro F1: Llama-3.1-8B: 0.337, Llama-3.3-70B: 0.325) due to extremely high FIR (0.73--0.75), while Qwen3-4B over-intervenes almost universally (FIR $= 0.897$). After fine-tuning, these models improve Macro F1 to 0.737--0.747, yielding gains ranging from $+0.400$ to $+0.531$ in Macro F1 over their respective zero-shot baselines. The largest gain in Macro F1 is observed for Qwen3-4B, which improves from 0.206 zero-shot to 0.737 after fine-tuning ($+0.531$), reinforcing that intervention timing is a learned capability, not an emergent property of pretraining scale.

The results displayed are of the SFT done on the \textit{Token Dataset}; \textit{Dialogue Dataset} results are in Appendix~\ref{app:tables}. Fine-tuning on When2Speak yields $+0.40$ to $+0.53$ Macro F1 absolute for capable model families, confirming that intervention timing is a learned, not emergent, capability.

\begin{table}[t]
\caption{SFT gain over same-model-family zero-shot baseline.}
\label{tab:sft}
\centering
\resizebox{\linewidth}{!}{%
\begin{tabular}{lcccccc}
\hline
\textbf{Model Family} & \textbf{0-shot Macro F1} & \textbf{SFT Macro F1} & \textbf{$\Delta$SFT Gain} & \textbf{0-shot FIR} & \textbf{SFT FIR} & \textbf{$\Delta$FIR} \\
\hline
GPT-4o-mini        & 0.569 & 0.697 & $+0.128$ & 0.257 & 0.163 & $-0.094$ \\
GPT-4.1-mini       & 0.499 & 0.729 & $+0.230$ & 0.435 & 0.105 & $-0.330$ \\
Qwen3-4B           & 0.206 & 0.737 & $+0.531$ & 0.897 & 0.037 & $-0.860$ \\
Llama-3.1-8B       & 0.337 & 0.740 & $+0.403$ & 0.731 & 0.044 & $-0.687$ \\
Qwen3-8B$^\dagger$ & 0.466 & 0.739 & $+0.273$ & 0.000 & 0.036 & $+0.036$ \\
Llama-3.3-70B      & 0.325 & 0.747 & $+0.422$ & 0.749 & 0.050 & $-0.699$ \\
\hline
\end{tabular}}
\par\smallskip
{\footnotesize $^\dagger$ Zero-shot collapses to always-SILENT (FIR\,=\,0.000); $\Delta$FIR is not meaningful.}
\end{table}

\subsection{Reinforcement Learning with Asymmetric Reward Shaping}

While supervised fine-tuning (SFT) improves intervention timing, it optimizes a symmetric objective that treats false interruptions and missed interventions equally. In practice, these errors carry asymmetric costs, leading SFT models to adopt overly conservative policies that prioritize avoiding interruptions at the expense of missing valuable interventions. To better capture this tradeoff, we employ reinforcement learning (RL), which enables direct optimization of cost-aware objectives and allows the model to learn a more balanced intervention strategy.

\subsubsection{Experimental Setup}

 We apply Group Relative Policy Optimization (GRPO) to the SFT-trained Llama-3.1-8B SFT, using LoRA (rank $= 16$, matching the SFT configuration to ensure comparability and avoid introducing additional hyperparameter variation between phases). We select the 8B model due to the negligible SFT performance gap with 70B ($+0.006$ F1) and lower computational cost.
 We evaluate sensitivity to reward weighting via a $\lambda$ ablation ($\lambda \in \{0.25, 0.50, 1.00\}$, 3 seeds) and isolate component contributions through a cumulative reward ablation.

\subsubsection{Reward Function}

We define a reward function that explicitly models the asymmetric costs of intervention timing (detailed description in Appendix~\ref{app:reward-function}):
\begin{equation}
R = R_{\text{accuracy}} + R_{\text{soft\_timing}} + R_{\text{type\_bonus}} - \lambda \cdot R_{\text{necessity}}
\end{equation}
The reward decomposes into four components. Accuracy provides a baseline symmetric signal, while necessity introduces an additional penalty on false interruptions, correcting the over-conservative bias observed under SFT. Type bonus reinforces contextually appropriate intervention types, and soft timing allows near-correct early interventions to receive partial credit.

Together, this formulation shifts learning from symmetric classification toward cost-aware intervention behavior, explicitly balancing over-intervention and under-intervention.

\subsubsection{Results}

Reinforcement learning improves intervention behavior by reducing Missed Intervention Rate (MIR) from 0.521 (SFT) to 0.186--0.218, while increasing SPEAK recall from 0.479 to 0.78--0.81. This demonstrates that RL overcomes the over-conservative bias of SFT, enabling the model to intervene more effectively. 
\begin{table}[t]
\caption{RL results (full reward, $\lambda=0.50$, $n=3$ seeds) compared to SFT baseline and zero-shot.}
\label{tab:rl_main}
\centering
\resizebox{\linewidth}{!}{%
\begin{tabular}{lcccc}
\hline
\textbf{Policy} & \textbf{Macro F1} & \textbf{SPEAK Recall} & \textbf{FIR $\downarrow$} & \textbf{MIR $\downarrow$} \\
\hline
Llama-3.1-8B zero-shot          & 0.337 & 0.872 & 0.731 & 0.128 \\
SFT Llama-3.1-8B (baseline)     & 0.740 & 0.479 & 0.044 & 0.521 \\
RL $\lambda=0.50$ ($n=3$ seeds) & $0.673 \pm 0.015$ & $0.814 \pm 0.024$ & $0.227 \pm 0.026$ & $0.186 \pm 0.024$ \\
\hline
\end{tabular}}
\end{table}

This improvement comes with an increase in False Interruption Rate (FIR) ($0.044 \to 0.195$--$0.227$), where increasing recall necessarily introduces additional false interventions. Across $\lambda$ values, performance is stable, with $\lambda = 0.50$ achieving the strongest MIR reduction. RL substantially reduces MIR at the cost of increased FIR, shifting the model from a conservative to an interventionist regime.

Compared to SFT, RL shifts the model from a conservative regime (high MIR, low FIR) to a more interventionist one (lower MIR, higher FIR). While Macro F1 decreases slightly ($-0.036$), the substantial recall gain ($+0.275$) indicates improved alignment with intervention opportunities. These results highlight that intervention standard SFT objectives and benefits from explicit cost-aware optimization do not fully capture intervention timinge and performance metrics for RL are in Appendix~\ref{app:SFTRL}.

\section{Societal Implications}

This work has the potential to improve human–AI interaction by enabling systems that participate in conversations more naturally and appropriately, reducing unnecessary interruptions and improving the usefulness of AI assistants in collaborative settings such as education, healthcare, and professional workflows. By modeling when to remain silent, such systems may also reduce cognitive overload and improve user trust.

At the same time, improved intervention capabilities may introduce risks if misapplied. Models that better infer when to speak could be used to strategically influence conversations, amplify persuasive messaging, or subtly insert misinformation in group settings. There are also fairness considerations, as intervention policies learned from synthetic or biased data distributions may not generalize equally across different communication styles or demographic groups. Additionally, deployment in sensitive domains raises privacy and safety concerns if models intervene inappropriately or at critical moments.

Mitigating these risks requires careful deployment practices, including monitoring intervention behavior, calibrating decision thresholds, and adapting models to domain-specific norms. We hope this work encourages further research on safe and responsible modeling of conversational participation.

\section{Limitations}

While When2Speak provides a scalable framework for modeling intervention timing, several directions remain for future work. First, the dataset is synthetically generated, and although grounded in real-world data, it may not fully capture the complexity of natural multi-party interactions. Second, our formulation focuses on turn-level decisions, leaving longer-term conversational dynamics and strategic timing unexplored. Third, the observed tradeoff between missed interventions and false interruptions under class imbalance suggests opportunities for improved calibration and objective design. Finally, extending the pipeline to domain-specific settings may require additional grounding and tailored intervention policies. We hope this work motivates further research on temporal participation and turn-taking in conversational AI.

\section{Conclusion}

We introduce When2Speak, a dataset and pipeline for learning intervention timing in multi-party conversations. Across models, fine-tuning on When2Speak yields substantial gains over zero-shot baselines, demonstrating that intervention timing is a learnable capability. Reinforcement learning further shifts model behavior from conservative to interventionist, reducing Missed Intervention Rate (MIR) from 0.521 to 0.186--0.218 while increasing recall from 0.479 to 0.78--0.81. Our results suggest that conversational intelligence extends beyond response generation to interaction-level decision-making, where deciding when to speak is as critical as what to say. By providing grounded supervision and a reproducible pipeline, When2Speak enables systematic study of this behavior across models and domains. This work points toward a new class of alignment problems centered on temporal participation and turn-taking, with implications for collaborative AI systems, human--AI facilitation, and multi-agent environments.

\section{Acknowledgment}
This work was supported in part by API credits from OpenAI through the Duke DeepTech AI for Metascience Initiative, and by Tinker API credits provided by Thinking Machines Lab.

\appendix

\appendix
\renewcommand{\thetable}{\Alph{section}.\arabic{table}}
\setcounter{table}{0}

\section{Prompts}
\label{app:prompts}

\subsection{Stage 2 Prompt}
\label{app:stage2-prompt}
 
\begin{quote}
\small
\texttt{You are a creative scenario writer. Your task is to generate a single, detailed scenario JSON object based on a user's question and its detailed background.}
 
\medskip
\texttt{Input Information:}\\
\texttt{1. Question Title: \{question\_title\}}\\
\texttt{2. Question Content: \{question\_content\}}\\
\texttt{3. Best Answer: \{best\_answer\}}
 
\medskip
\texttt{Task: Based on the provided information, create a complete scenario by performing these steps:}\\
\texttt{1. Invent a Social Context: Create a one-sentence context describing who would be discussing this topic.}\\
\texttt{2. Select an Intervention Type: Choose the most logical ai\_intervention\_type from: [Factual Correction, Concept Definition, Data Provision, Source Identification, Synthesis \& Reframing].}
 
\medskip
\texttt{Output Format: You must output ONLY the raw JSON object with the following structure:}\\
\texttt{\{\ "social\_context": "your one-sentence context here",\ "intervention\_type": "selected intervention type here"\ \}}
\end{quote}

\subsection{Stage 2 Options}
\label{app:stage2-option}

\begin{quote}
\small
\textbf{conversation\_styles:}
``with frequent disagreements and debates'',
``with participants building on each other's ideas collaboratively'',
``with a mix of experts and novices asking clarifying questions'',
``with storytelling and personal anecdotes'',
``with structured turn-taking and formal language'',
``with casual, overlapping dialogue and interruptions''

\textbf{tone\_variations:}
``enthusiastic and energetic'',
``thoughtful and contemplative'',
``professional and business-like'',
``casual and friendly'',
``curious and inquisitive''

\textbf{ask\_follow\_up:}
``ask a follow up question, to which [AGENT] should respond appropriately.'',
``refute what [AGENT] is saying, to which [AGENT] should give a deeper explanation.'',
``agree to what [AGENT] is saying, acknowledge it, and move on with the conversation.'',
``completely disregard what [AGENT] says and move on with the conversation.''
\end{quote}

\subsection{Stage 3 Transcript Generation Prompt}
\label{app:stage3-prompt}

\begin{quote}
\small
\texttt{You are a sophisticated data generator. Your task is to generate a realistic group discussion transcript based on the provided scenario.}

\texttt{Rules:}\\
\texttt{1. The discussion must feature \{human\_count\} human participants and one AI assistant named [AGENT].}\\
\texttt{2. [AGENT] appears only a few times (1--3 times), with a meaningful intervention.}\\
\texttt{3. The discussion should feel natural, with a clear trigger for [AGENT]'s intervention.}\\
\texttt{4. After [AGENT] speaks, humans should react naturally and continue the discussion.}\\
\texttt{5. Generate a relatively decent-sized conversation (at least 20 exchanges total).}\\
\texttt{6. Make the conversation \{selected\_style\} and maintain a \{selected\_tone\} tone.}\\
\texttt{7. Each participant should have a distinct personality and speaking style.}\\
\texttt{8. Output only the conversation, DO NOT OUTPUT ANYTHING BUT THE CONVERSATION.}\\
\texttt{9. After every human speaking, if the AI is not adding something meaningful to the conversation, add --- [AGENT]: >}\\
\texttt{10. Make [AGENT]'s intervention natural; do not include numbered lists or bullet points.}\\
\texttt{11. When the user asks a question to [AGENT] directly, and it responds, the user should \{follow\_up\}}\\
\texttt{12. The participants should treat [AGENT] like their AI partner in the conversation.}\\
\texttt{13. There should be no punctuation marks in your output.}\\
\texttt{14. Do not use names; use Speaker\_0, Speaker\_1, etc.}\\
\texttt{15. Do not have the users address [AGENT] explicitly unless needed in the follow-up.}

\texttt{Scenario Details:}\\
\texttt{- Topic: \{question\_title\}}\\
\texttt{- Initial Question: \{question\_content\}}\\
\texttt{- Context: \{context\}}\\
\texttt{- AI Intervention Type: \{intervention\_type\}}\\
\texttt{- Reference Answer (for context): \{best\_answer\}}
\end{quote}

\subsection{Zero-Shot Evaluation System Prompt}
\label{app:zero-shot-prompt}

\begin{quote}
\small
\texttt{You are [AGENT], an AI conversational agent participating in a multi-party discussion. Your role is to contribute meaningfully when appropriate, but also to exercise restraint --- remaining silent when others are engaged in dialogue that does not require your input.}

\texttt{At each turn, you will see the recent conversation history. Decide whether to speak or stay silent.}

\texttt{Respond with exactly one character:}\\
\texttt{< if you should speak at this turn (you have something valuable to add, someone needs help, or there is a natural opening)}\\
\texttt{> if you should remain silent (the conversation is flowing well without you, or your input is not needed)}

\texttt{Output only < or > --- no explanation, no other text.}
\end{quote}

\section{Dialogue Dataset Results}
\label{app:tables}

\begin{table}[h]
\caption{\textit{Dialogue Dataset} SFT results (test set, $n=21{,}687$). ROUGE-L was computed only on true-positive turns where the model correctly decided to speak. All models suffer substantial MIR increase relative to the \textit{Token Dataset} task.}
\label{tab:dialogue_sft}
\centering
\resizebox{\linewidth}{!}{%
\begin{tabular}{lcrrrrc}
\toprule
\textbf{Model} & \textbf{Params} & \textbf{Macro F1} & \textbf{Recall} & \textbf{FIR $\downarrow$} & \textbf{MIR $\downarrow$} & \textbf{ROUGE-L (TP)} \\
\midrule
Qwen3-4B-Instruct     & 4B  & 0.6398 & 0.2111 & 0.0058 & 0.7889 & 0.277 \\
Qwen3-8B              & 8B  & 0.6501 & 0.2279 & 0.0069 & 0.7721 & 0.271 \\
Llama-3.1-8B-Instruct & 8B  & 0.6527 & 0.2322 & 0.0071 & 0.7678 & 0.273 \\
Llama-3.3-70B-Instruct& 70B & 0.6673 & 0.2569 & 0.0085 & 0.7431 & 0.267 \\
\bottomrule
\end{tabular}}
\end{table}

\section{Reinforcement Learning with Asymmetric Reward Shaping}
\label{app:rl}

\subsection{Experimental Setup}

\paragraph{Algorithm.} We apply Group Relative Policy Optimization (GRPO; \citet{shao2024}) to the Llama-3.1-8B SFT checkpoint. GRPO generates $G=8$ rollouts per prompt, computes the group-normalised advantage --- (reward $-$ group mean) / group std --- and backpropagates through a LoRA adapter of the same rank$=16$ configuration used during SFT. We select Llama-3.1-8B over Llama-3.3-70B because the SFT performance gap between these model sizes is only $+0.006$ Macro F1, while the 70B model incurs approximately $8\times$ the per-step compute cost.

\paragraph{Training Details.} Each training run consists of $500\ \text{batches} \times 32\ \text{prompts} \times 8\ \text{rollouts} = 128{,}000$ total forward passes; 500 batches was chosen based on pilot runs showing that training-batch reward plateaued at $\sim$400 steps, with 500 providing a buffer for late-stage adjustment. The group size $G=8$ follows the GRPO default used in \citet{shao2024}, providing sufficient rollout variance for stable advantage normalisation. We use a learning rate of $5\times10^{-6}$ with no warm-up. Critically, batches are sampled with a balanced 50\% SPEAK / 50\% SILENT constraint. Without this constraint, randomly sampled batches contain approximately 28 SILENT and 4 SPEAK examples, and the policy immediately collapses to always-SILENT: silence on SILENT turns yields zero reward with no corrective gradient. Balanced sampling prevents this collapse while preserving training signal on both classes.

\paragraph{Ablations.} We conduct two ablations. The lambda ablation runs 9 independent training sessions: three $\lambda$ values $(0.25, 0.50, 1.00) \times$ three random seeds $(42, 123, 7)$. The component ablation runs four cumulative reward configurations at $\lambda=0.50$, seed$=42$, adding one reward component at a time to isolate each contribution.
 The $\lambda$ range spans one octave below and above the midpoint $0.5$, providing a symmetric coarse sweep of the FIR penalty strength; three seeds per $\lambda$ are used to estimate training stochasticity without exceeding the available compute budget.

\subsection{Reward Function}
\label{app:reward-function}

The total reward $R = R_{\text{accuracy}} + R_{\text{soft\_timing}} + R_{\text{type\_bonus}} - \lambda \cdot R_{\text{necessity}}$ decomposes into four components designed to encode the asymmetric cost structure of the intervention timing problem. Table~\ref{tab:reward} summarises each component.

The design rationale is as follows. Standard cross-entropy assigns equal loss to false interruptions and missed interventions, producing the over-conservative SFT policy documented in Section~5.1. $R_{\text{accuracy}}$ replicates this symmetric signal within the RL reward, providing a stable baseline gradient. $R_{\text{necessity}}$ breaks the symmetry: a $-\lambda$ penalty applied to every false interruption, independently of $R_{\text{accuracy}}$, raises the cost of over-speaking relative to remaining silent without the corresponding gradient pulling the policy back toward silence on missed turns. $R_{\text{type\_bonus}}$ teaches the model which type of epistemic gap motivates an intervention --- Synthesis, Factual Correction, and so on --- improving precision without directly sacrificing recall. $R_{\text{soft\_timing}}$ encodes the practical tolerance for slightly early synthesis: intervening one turn before the gold label is more useful to the conversation than staying silent entirely, and rewarding near-correct timing reduces penalisation of temporally proximate interventions.

\begin{table}[h]
\caption{Reward function components. $\lambda \in \{0.25, 0.50, 1.00\}$ is the main ablation hyperparameter controlling the FIR/MIR trade-off. $R_{\text{type\_bonus}}$ uses GPT-4o-mini as a judge to classify intervention type on true-positive turns.}
\label{tab:reward}
\centering
\resizebox{\linewidth}{!}{%
\begin{tabular}{llll}
\toprule
\textbf{Component} & \textbf{Value} & \textbf{Applied when} & \textbf{Purpose} \\
\midrule
$R_{\text{accuracy}}$     & $+1$ correct, $-1$ wrong & Always & Core binary timing signal \\
$R_{\text{soft\_timing}}$ & $+0.1$ ($\leq 2$ turns early), $0.0$ ($\leq 4$), $-0.1$ ($>4$) & SPEAK turns & Tolerates slightly early interventions \\
$R_{\text{type\_bonus}}$  & $+0.3$ if type judgment matches gold & True positives only & Rewards knowing why to intervene \\
$R_{\text{necessity}}$    & $-\lambda$ penalty on false interruptions & False positives only & Encodes asymmetric social cost \\
\bottomrule
\end{tabular}}
\end{table}

\subsection{Lambda Ablation Results}

Table~\ref{tab:lambda} reports mean $\pm$ std across three seeds for each $\lambda$ value, evaluated on the full 21,687-example test set. The SFT Llama-3.1-8B baseline and Llama zero-shot are included for reference.

\begin{table}[h]
\caption{Lambda ablation results (mean $\pm$ std, $n=3$ seeds per $\lambda$). All RL policies were evaluated on the full 21,687-example test set. $\Delta$MIR $=$ absolute reduction in Missed Intervention Rate relative to the SFT baseline.}
\label{tab:lambda}
\centering
\resizebox{\linewidth}{!}{%
\begin{tabular}{lccccc}
\toprule
\textbf{Policy} & \textbf{Macro F1} & \textbf{SPEAK Recall} & \textbf{FIR $\downarrow$} & \textbf{MIR $\downarrow$} & \textbf{$\Delta$MIR vs SFT} \\
\midrule
Llama-3.1-8B zero-shot          & 0.337 & 0.872 & 0.731 & 0.128 & --- \\
SFT Llama-3.1-8B (baseline)     & 0.740 & 0.479 & 0.044 & 0.521 & --- \\
RL $\lambda=0.25$ ($n=3$ seeds) & $0.691 \pm 0.007$ & $0.782 \pm 0.017$ & $0.195 \pm 0.013$ & $0.218 \pm 0.017$ & $-0.303$ \\
RL $\lambda=0.50$ ($n=3$ seeds) & $0.673 \pm 0.015$ & $0.814 \pm 0.024$ & $0.227 \pm 0.026$ & $0.186 \pm 0.024$ & $-0.335$ \\
RL $\lambda=1.00$ ($n=3$ seeds) & $0.681 \pm 0.008$ & $0.808 \pm 0.015$ & $0.214 \pm 0.014$ & $0.192 \pm 0.015$ & $-0.329$ \\
\bottomrule
\end{tabular}}
\end{table}

All three RL policies dramatically reduce MIR relative to the SFT baseline. MIR falls from 0.521 to 0.186--0.218 --- a 0.30--0.34 absolute reduction --- and SPEAK recall rises from 0.479 to 0.78--0.81, shifting the model from the SFT regime of ``when in doubt, stay silent'' to an actively interventionist policy. This MIR improvement is the central result of the RL stage: it directly addresses the structural ceiling that SFT cannot overcome through scaling alone.

FIR rises correspondingly from 0.044 (SFT) to 0.195--0.227, exceeding the deployment target of FIR $\leq 0.10$. Two mechanisms drive this. First, there is a distributional mismatch between training and evaluation: batches are 50\% SPEAK during training, but the test set is 87\% SILENT, so the learned policy is calibrated for a less skewed environment than it is evaluated in. Second, there is a structural recall--precision coupling at high class imbalance: recovering the $\sim$925 additional warranted interventions represented by the MIR reduction requires the policy to adopt a more liberal speak threshold, but at 87\% SILENT there are $6.8\times$ more silence opportunities per warranted intervention --- this asymmetry geometrically amplifies false interruptions even under modest threshold liberalisation. These two effects cannot be fully decoupled through reward shaping alone. Decision-threshold calibration at inference time --- sweeping the logprob threshold for the speak token on a held-out development set --- is the natural next step for finding an operating point that satisfies the FIR constraint while preserving the recall gains from RL training.

The expected monotonic pattern (higher $\lambda \to$ lower FIR, higher MIR) holds approximately across $\lambda$ values but not cleanly across seeds. $\lambda=0.25$ achieves FIR $= 0.195$ and MIR $= 0.218$; $\lambda=0.50$ achieves FIR $= 0.227$ and MIR $= 0.186$; $\lambda=1.00$ achieves FIR $= 0.214$ and MIR $= 0.192$. Seed variance is substantial, particularly at $\lambda=0.50$ (FIR std $= 0.032$), reflecting that training-batch FIR does not reliably predict test-set FIR because the sampling distributions differ by design. Within this variance, $\lambda=0.50$ achieves the strongest MIR reduction ($\Delta$MIR $= -0.335$).

\subsection{Reward Component Ablation}

Table~\ref{tab:component} reports the cumulative effect of adding each reward component, evaluated at $\lambda=0.50$, seed$=42$. Each row adds one component to the previous configuration.

\begin{table}[h]
\caption{Cumulative reward component ablation ($\lambda=0.50$, seed$=42$). $R_{\text{type\_bonus}}$ is the single highest-impact component, contributing $+0.016$ F1 and $-0.039$ FIR.}
\label{tab:component}
\centering
\resizebox{\linewidth}{!}{%
\begin{tabular}{lcccccc}
\toprule
\textbf{Reward Configuration} & \textbf{Macro F1} & \textbf{SPEAK Recall} & \textbf{FIR $\downarrow$} & \textbf{MIR $\downarrow$} & \textbf{$\Delta$F1} & \textbf{$\Delta$FIR} \\
\midrule
$R_{\text{accuracy}}$ only          & 0.680 & 0.799 & 0.213 & 0.201 & ---    & ---    \\
$+\ R_{\text{soft\_timing}}$         & 0.688 & 0.799 & 0.204 & 0.201 & $+0.008$ & $-0.009$ \\
$+\ R_{\text{type\_bonus}}$          & 0.704 & 0.738 & 0.165 & 0.262 & $+0.016$ & $-0.039$ \\
Full reward ($+\ R_{\text{necessity}}$) & 0.704 & 0.754 & 0.170 & 0.246 & $\approx 0$ & $-0.016$ MIR \\
\bottomrule
\end{tabular}}
\end{table}

$R_{\text{type\_bonus}}$ is the most impactful single component, yielding $+0.016$ Macro F1 and $-0.039$ FIR --- the largest step-gain in the ablation. Teaching the model, which type of epistemic gap motivates an intervention improves selectivity: the model becomes less likely to falsely interrupt turns that do not correspond to a recognisable intervention pattern, reducing over-speaking without a direct penalty on false positives. This precision gain comes at the cost of recall (SPEAK recall drops from 0.799 to 0.738), reflecting the expected precision--recall trade-off when the policy becomes more selective. Adding the full $R_{\text{necessity}}$ penalty partially recovers recall to 0.754 while further reducing MIR by 0.016 with negligible F1 cost, confirming that the two components are complementary rather than redundant. $R_{\text{soft\_timing}}$ contributes a smaller $+0.008$ F1 and $-0.009$ FIR improvement by allowing the model to receive positive reward for near-correct early interventions rather than treating all off-gold-turn predictions as equally incorrect.

\subsection{SFT vs.\ RL: Summary}
\label{app:SFTRL}

Table~\ref{tab:sft_vs_rl} summarises the net shift from the SFT baseline to the best-performing RL configuration (full reward, $\lambda=0.50$).

\begin{table}[h]
\caption{SFT vs.\ RL (full reward, $\lambda=0.50$) on the 21,687-example test set. RL trades a moderate Macro F1 drop and higher FIR for a large MIR reduction and recall gain. DA/NI decomposition uses the same slices as Table~2.}
\label{tab:sft_vs_rl}
\centering
\resizebox{\linewidth}{!}{%
\begin{tabular}{lccc}
\toprule
\textbf{Metric} & \textbf{SFT Llama-3.1-8B} & \textbf{RL Full ($\lambda=0.50$)} & \textbf{Change} \\
\midrule
Macro F1      & 0.740 & 0.704 & $-0.036$ \\
SPEAK Recall  & 0.479 & 0.754 & $+0.275$ \\
FIR           & 0.044 & 0.170 & $+0.126$ \\
MIR           & 0.521 & 0.246 & $-0.275$ \\
DA Macro F1   & 0.850 & 0.823 & $-0.027$ \\
NI Macro F1   & 0.707 & 0.677 & $-0.030$ \\
\bottomrule
\end{tabular}}
\end{table}

RL with asymmetric reward shaping successfully repositions the operating point from SFT's over-conservative regime (FIR $= 0.044$, MIR $= 0.521$) to an actively interventionist one (FIR $= 0.170$, MIR $= 0.246$). Macro F1 falls slightly ($-0.036$) because F1 penalises both precision and recall losses symmetrically: RL gains substantial recall ($+0.275$) but incurs a precision penalty from elevated FIR, and these effects partially cancel in the F1 aggregate. In facilitation settings where missing a warranted intervention is more disruptive than occasional over-speaking, the RL policy is strictly preferable to SFT. The DA/NI gap is largely preserved ($\Delta = 0.143$ under SFT vs.\ $\Delta = 0.146$ under RL), indicating that RL has not introduced new dependence on direct-address cues; contextual no-invitation reasoning remains an open challenge. The remaining FIR gap above the $\leq 0.10$ deployment target appears to arise from the interaction between strong class imbalance and balanced-batch training, and is more effectively addressed through inference-time log-probability threshold calibration than additional reward shaping.

\end{document}